# Exploiting Context to Identify Lexical Atoms
## -- A Statistical View of Linguistic Context

Chengxiang Zhai

Laboratory for Computational Linguistics
Carnegie Mellon University
Pittsburgh, PA 15213
U.S.A.
*Email: cz25@andrew.cmu.edu*

**Abstract**

Interpretation of natural language is inherently context-sensitive. Most words in natural language are ambiguous and their meanings are heavily dependent on the linguistic context in which they are used. The study of lexical semantics can not be separated from the notion of context. This paper takes a contextual approach to lexical semantics and studies the linguistic context of lexical atoms, or "sticky" phrases such as "hot dog". Since such lexical atoms may occur frequently in unrestricted natural language text, recognizing them is crucial for understanding naturally-occurring text. The paper proposes several heuristic approaches to exploiting the linguistic context to identify lexical atoms from arbitrary natural language text.

## 1. Introduction

Human communication relies heavily on the mutual understanding of the context or situation where the communication is performed. It is not a surprise that interpretation of natural language is inherently context-sensitive. In different situations or contexts, the same sentence may be interpreted differently; anaphors may be resolved differently; structural and lexical ambiguity may be resolved in different ways. Thus, context, because of its importance to natural language understanding, has been a very important topic studied by computational linguists[Allen 95, Alshawi 87].

The importance of context for lexical semantics has been emphasized[Cruse 86, Rieger 91, Slator 91, Schutze 92]. Most words in natural language are ambiguous, and may be interpreted differently within different contexts. For example, "bank" can mean an institution that looks after your money (the "money" sense) or the side of a river (the "river" sense). To decide which sense "bank" takes in an actual situation we must look at the context of "bank". Sometimes a small amount of information (such as in the phrase "high interest bank") is sufficient for the disambiguation; while in other cases, a larger amount of information may be needed, (E.g., in the sentence "He went to the bank yesterday", the sentence itself is insufficient for the disambiguation and a richer context is needed.) The context of a word can usually provide good clues to decide which sense the word has in this context, or as Firth said, "You shall know a word by the company it keeps" [Firth 57]. Thus, if "money", "account", or "interest" occurs in the context, it is very likely "bank" has the "money" sense; while if "river", or "water" occurs in the context, it is more likely to take the "river" sense.

In this paper, we study the linguistic context of a special kind of lexical units called lexical atoms. A lexical atom is a "sticky" phrase like "hot dog", in which one or more constituent words do not carry their regular meanings. Since lexical atoms are multi-word lexical units that can not be processed compositionally, recognizing them is crucial for many natural language processing tasks. We propose several statistical heuristics to exploit the context to identify lexical atoms from unrestricted natural language text.

cmp-lg/9701001  2 Jan 1997

## 2. Lexical Acquisition and Lexical Atoms

The study of lexical semantics and lexicon acquisition have been given much attention recently by computational linguists [Zampolli et al 95, Saint-Dizier 95, Zernik 91]. One reason for this interest may be due to the fact that many modern grammar theories are now converging on the acceptance of the increasingly important role of lexicon and a theory of lexicon is a very important part of the grammar theory. Another reason may be due to the fact that the availability of a large-scale lexicon is necessary for the scale-up of any practical natural language processing system. There are two different approaches to developing a lexicon: manual construction and automatic acquisition. Manual construction of lexicon is both time-consuming and labor-intensive, automatically acquiring a lexicon is thus very attractive [Zernik 91].

One important aspect of acquiring a lexicon is the acquisition of lexical atoms. A lexical atom is a multiple-word phrase that has a non-compositional meaning, that is, the meaning of the phrase is not a direct composition of the literal meaning of the individual words that comprise the phrase. A good example is "hot dog", where the meaning of the whole phrase has almost nothing to do with the literal meaning of "hot" or "dog". Proper names and some technical terms are also good examples (e.g., "Hong Kong", "artificial intelligence"). New phrases that people constantly invent are often typical lexical atoms (e.g., "TV dinner").

Because the meaning of lexical atoms is non-compositional, naturally, they must be recognized and treated as a single unit, rather than a composite structure. Lexicographists have to identify lexical atoms and list them as independent lexicon entries[Hartmann 83]. In information retrieval, it is desirable to recognize and use lexical atoms for indexing[Evans & Zhai 96]. In machine translation, a lexical atom needs to be translated as a single unit, rather than word by word [Meyer et al 90].

The general issue of the compositionality of meaning has been widely studied by linguists and computational linguists[See e.g., Dowty et al 81, Pustejovsky et al 92, Pereira 89, Pereira & Pollack 90, Dale 89 among others]. One difficulty with lexical atoms is that they are inherently context-sensitive. For example, "White House" is a lexical atom in a government news report, but may not be a lexical atom in a general news report (such as in "a big white house"). Thus, it is natural to exploit context to identify lexical atoms.

## 3. Exploiting context to identify lexical atoms

The study of lexical semantics can take two major approaches -- the generative approach [Pustekpvslu 95] or the contextual approach [Cruse 86, Evens 88]. In the generative approach, a lexical item is *defined* based on certain more basic notions (such as conceptual primitives). The intensional and extensional content of the lexical item is described; In the contextual approach, a lexical item is *related* to other lexical items. The relation or dependency among lexical items is described. For lexical acquisition, the generative approach often exploits online dictionaries and extracts lexical information from the entry definition; while the contextual approach often exploits the online text or large corpora to extract lexical collocations. Because the meaning of a lexical atom has a non-compositional nature, it is inherently hard to identify lexical atoms using the generative approach. However, if we take the contextual view, and regard the meaning of any phrase as being "defined" by the context in which the phrase is used, it is possible to exploit such context to decide if a phrase is likely to be a lexical atom.

As mentioned above, the most important characteristic of a lexical atom is its semantic non-compositionality, that is, the meaning of a lexical atom is different from any simple combination of the normal literal meanings of its component words. In other words, not every individual words keeps its normal literal meaning. Thus, we may define a two-word lexical atom roughly as follows. Such definition can be generalized to lexical atoms with more than two words, but we are only concerned with two words in this paper.

**Lexical Atom**: A two word noun phrase [X Y] is a lexical atom, if and only if the meaning of [X Y] is not a direct composition of the regular literal meanings of X and Y.

For instance, "natural gas" (as in the name of a gas company) can be judged as a lexical atom, because the meaning of "natural gas" is not a direct combination of the meaning of "natural" and the meaning of "gas". Although "gas" seems to have its regular meaning here, "natural" does not carry its regular meaning at all, and thus, "natural gas" is not "natural". On the other hand, "stock market" would not be judged as a lexical atom, because both "stock" and "market" carry their regular meanings in contributing to the meaning of the whole phrase "stock market". Note that the definition above implies that the judgment of lexical atom is context-dependent, as the judgment is based on the phrase's meaning, which is context-dependent. For example, whether "white house" is a lexical atom has to be judged based on the context in which it occurs (see section 2). Even with the context given, whether the meaning of a phrase is compositional can still be uncertain, since the notion of compositionality is not clear-cut. For example, it is hard to say that the phrase "bottom line" has a compositional or non-compositional meaning. Thus, in order to judge and identify lexical atoms, it is necessary to measure the composition uncertainty in a context-sensitive way.

The linguistic context of a lexical atom may be represented with different levels of details. Theoretically, it is desirable to exploit as much contextual information as possible, but in practice, it is extremely hard to obtain a deep and detailed representation of context in unrestricted domains, because even to correctly parse an unrestricted text is very hard. For this reason, an extremely simplified view of context, where a context is simply represented by the words that occur in the context, has been common in corpus-based linguistics and statistical natural language processing. This view is especially common in the work of word sense disambiguation, where a fixed length of word window (or context) is used to characterize the meaning or sense of the word in the center [Schutze 92, Gale et al 92]. For the purpose of lexical atom identification, we follow the same kind of simplification. Specifically, we restrict the linguistic context of a word/phrase to the surrounding words of the word/phrase within a certain boundary (e.g., a 50-word, or 10-sentence text window). This, of course, has the disadvantage of excluding not only the pragmatic environment (e.g., the language users), but also any syntactic relation, from the notion of context. Nevertheless, the simplification makes it much easier to use the contextual information, especially by statistical approaches.

With this simplified notion of context, we propose three statistical heuristics to measure the compositionality:

- **Phrase-Word Cooccurrence (PWC) measure**: The compositionality of a phrase [X Y] is proportional to the relative frequency of X's (or Y's) occurring in the context of [X Y].

This measure is based on the observation that if the meaning of [X Y] is radically different from X (or Y), X (or Y) will be unlikely to occur in the context of [X Y]. For example, we would expect "hot" or "dog" to occur less frequently in the context of "hot dog" than "stock" or "market" would occur in the context of "stock market".

- **Word Association (WA) measure:** The compositionality of a phrase [X Y] is proportional to the statistical association of X and Y in other ways than the phrase [X Y] within some boundary of context.

This measure is based on the observation that if [X Y] is non-compositional, then, it is unlikely that the association [X Y] comes from the general semantic association between X and Y. Thus, if we take the association [X Y] away from the text, we will not expect X and Y to have strong association. For example, we would not expect "hot" and "dog" to have strong associations other than the one in "hot dog", but "stock" and "market" may still be associated in contexts other than the phrase "stock market".

- **Context Similarity (CS) measure**: The compositionality of a phrase [X Y] is proportional to the similarity of the context of [X Y] and that of X or Y.

This measure is based on the observation that if [X Y] is non-compositional, we will expect the context of [X Y] and that of X or Y to be very different. For example, we will expect the context of "hot dog" to be very different from the general context of the single word "hot" or "dog", but "stock market" has a much better chance to share certain context with the single word "stock" or "market".

Based on these heuristics, it is possible to quantitatively measure the compositionality of a phrase using simple statistics. For example, the following is one of the possible ways to compute the three measures.

We assume that the context of any word/phrase is just the surrounding N words of the word/phrase. Let [X,Y] denote a phrase with words X and Y; let FQ(X) be the total frequency of X in the corpus, and FQ(X:Y) the total frequency that X and Y co-occur within an N-word text window.

The PWC measure of a phrase [u,v] (with respect to u and v respectively) can be computed as

$$PWC([u,v],u) = \frac{FQ(u:[u,v])}{FQ([u,v])}$$

$$PWC([u,v],v) = \frac{FQ(v:[u,v])}{FQ([u,v])}$$

The WA measure of a phrase can be computed as

$$WA([u,v]) = \frac{FQ(u:v) - FQ([u,v])}{FQ([u,v])}$$

To compute the CS measure of a phrase [u,v], we need first to quantitatively represent the context of a word. This can be done by forming a context vector for any word using the same method as [Schutze 92], where the vector was used for the task of word sense disambiguation. The basic idea is the following: Given a word w, we can collect all the context words (i.e., words that occur in the surrounding N-word contexts of w). Each context word is then given a weight that measures its association with w, and the weights of all the context words of w form a vector that quantitatively characterizes the context of w. The weight of a context word can be determined simply by counting the occurrences of the word in all the N-word contexts of w, as in [Schutze 92], or also considering the global frequency of the context word. The latter is similar to the idf-tf scoring in standard information retrieval techniques [Salton & McGill 83]. With the idf-tf scoring, the weight of a context word c is proportional to its frequency in the N-word contexts of w, but inversely proportional to its global frequency, and can be computed as "FQ(c:w)/log(FQ(c)+1)". Since the contexts of u and v can be represented by two vectors of numbers, the similarity of their contexts can be measured by any standard similarity measure between two vectors. For example, given two vectors $v1=(x1,x2,...xn)$ and $v2=(y1,y2,...,yn)$, the cosine measure of the similarity between v1 and v2 is

$$sim(v1,v2) = \frac{x1*y1 + x2*y2 +... xn*yn}{sqrt(x1*x1+...+xn*xn)*sqrt(y1*y1+...+yn*yn)}$$

For the WA measure, we get one single compositionality value, but for the PWC and CS measures, we get two values with respect to each word. To obtain one single combined compositionality value for the PWC and CS, we can take the sum or product of the two individual values. Once the compositionality of a phrase can be quantitatively measured, lexical atoms may be identified by selecting, as lexical atoms, those phrases whose compositionality measure value is below certain threshold.

## 4. Experiments

To test our heuristics for measuring phrase compositionality and identifying lexical atoms from naturally-occurring text, we implemented the three measures of phrase compositionality and performed some experiments on an experiment corpus -- the initial 20-megabyte chunk from the Associate Press Newswire 89 corpus (from the Tipster Collection [Harman 95]). In all experiments, the experiment corpus was first tagged with a lexicon. Then all the functional words were dropped and the corpus was thus condensed to contain only substantial words (mainly nouns, verbs, adjectives)[1]. The 400 most frequent noun-noun pairs in the corpus (with a frequency range of 11-700) were then generated as the candidate set of lexical atoms.

Each of the three measures was used to generate a ranked list of the 400 candidates based on their compositionality measure values. For the PWC measure, the combined compositionality value is the product of the two PWC values with respect to each word [2]. For the CS measure, the combined CS measure is the sum of two individual CS values, and the idf-tf scoring is used. All experiments, unless otherwise specified, used the same window size 80, i.e., the preceding 40 words and the following 40 words of any candidate phrase are regarded as the "context" of the phrase.

Since non-compositional pairs are often also collocations, it makes sense to see if the measures for finding collocations can also capture compositionality. Several such measures have been proposed [See for example, Smadja et al 90, Church et al 89, and Dunning 93]. We tried one very popular measure that has been frequently used for finding collocations -- Mutual Information (MI). Informally, the MI measure of any two words compares the probability of observing the two words together (as an adjacent pair) with the probabilities of observing the two words independently. A greater MI value indicates that the two words occur as an adjacently pair much more frequently than the expected chance. Details of this measure can be found in [Church et al 89].

The top 10 and bottom 5 word pairs from each measure are shown in Table 1 and Table 2 respectively. It is easy to see that most top candidates are indeed good lexical atoms, while the bottom ones are generally not, which indicates that all the four measures can capture the compositionality to certain extent.

| PWC measure | WA measure | CS measure | MI measure |
|---|---|---|---|
| asylum seeker | affirmative action | blue chip | pork belly |
| black tie | asylum seeker | Savannah River | asylum seeker |
| blue chip | blue chip | square root | Italian lira |
| death penalty | drug czar | enforcement official | feeder cattle |
| drug czar | en route | red cross | en route |
| en route | Holly Farm | guest house | Swiss franc |
| feeder cattle | Italian lira | French franc | Warsaw Pact |
| Holly Farm | Jimmy Carter | Naturalization Service | Jimmy Carter |
| independent counsel | jumbo jet | operation rescue | wind chill |
| Jimmy Carter | Naturalization Service | surgeon general | Savannah River |

**Table 1. The top 10 word pairs for four different measures**

---

[1] The size of the condensed corpus is around 14 megabytes.
[2] Ranking by the sum of the two PWC values seemed to work less well than ranking by the product.

| PWC measure | WA measure | CS measure | MI measure |
|---|---|---|---|
| school student | police official | income tax | police official |
| reagan budget | company official | pacific coast | state official |
| dollar rate | state official | tender offer | state police |
| african student | government report | African student | last time |
| Chinese student | school student | aids virus | last year |

**Table 2. The bottom 5 word pairs for four different measures**

In order to quantitatively compare each measure, the author manually judged 152 of the 400 candidates as true lexical atoms. It is difficult to use any well-defined criterion for the judgment, but generally, a pair would be judged as a lexical atom if the description of the pair's meaning does not involve the constituent words or the relation between the two words is so complex that extra substantial words are needed to describe it. For example, because "stock market" can be described as the "market" of "stock" and "Japanese student" can be described as the "student" who is "Japanese", neither "stock market" nor "Japanese student" has been judged as a lexical atom. But "real estate" has been judged as a lexical atom because "real estate" is not an "estate" that is "real". Although there are many vague cases and we can even imagine a great inconsistency between human judgments, there are also many clear cases, and the judgment may still be useful for the purpose of comparison of different measures.

Each measure is evaluated by the precisions at 11 selected points, each of which corresponds to a pre-specified cutoff of the ranked candidate pairs. For example, if among the top 10 pairs, 8 of them are true lexical atoms, we have a precision of 0.8 at 10 candidates (at a cutoff of 10). We also calculated the average precision, which is the average (over all the lexical atoms) of the precisions at all points of finding each new lexical atom.

Table 3 compares the four measures, and Figure 1 shows the corresponding precision curves.

|  | PWC measure | WA measure | CS measure | MI measure |
|---|---|---|---|---|
| **at 10 pairs** | 0.900 | 0.800 | 0.500 | 0.700 |
| **at 20 pairs** | 0.900 | 0.850 | 0.600 | 0.850 |
| **at 30 pairs** | 0.900 | 0.833 | 0.600 | 0.833 |
| **at 40 pairs** | 0.875 | 0.825 | 0.625 | 0.825 |
| **at 50 pairs** | 0.820 | 0.740 | 0.600 | 0.820 |
| **at 70 pairs** | 0.757 | 0.729 | 0.586 | 0.771 |
| **at 100 pairs** | 0.710 | 0.700 | 0.540 | 0.680 |
| **at 150 pairs** | 0.633 | 0.647 | 0.500 | 0.613 |
| **at 200 pairs** | 0.535 | 0.580 | 0.485 | 0.565 |
| **at 300 pairs** | 0.470 | 0.467 | 0.420 | 0.457 |
| **at 400 pairs** | 0.380 | 0.380 | 0.380 | 0.380 |
| **Average** | **0.681** | **0.678** | **0.515** | **0.655** |

**Table 3. Precisions for four different measures**

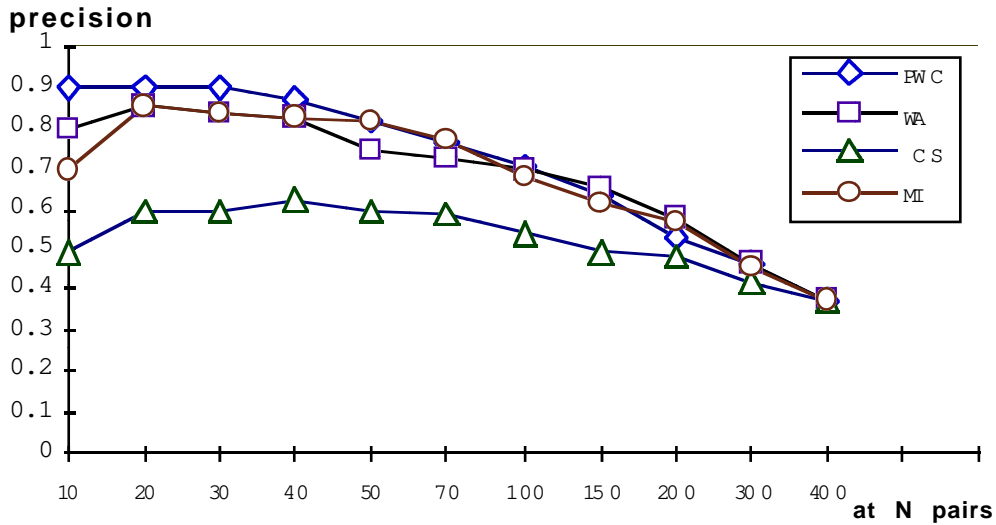

**Figure 1. Precision curves for four different measures**

From these results, we can see that the PWC measure gives the highest precision, especially in terms of the front-end precision. The front-end precision is a good indicator of how well a measure captures compositionality, since a higher front-end precision indicates that the measure has "pulled" up more true lexical atoms on the top. The shape of the precision curve can also reflect how accurate a measure is. For example, the shape of the PWC curve tells us that we get a greater ratio of false lexical atoms as the threshold for selecting lexical atoms is being relaxed (i.e., as the word pair cutoff increases). This is what we may expect for an accurate measure of lexical atoms. However, for all the other three measures, we may get a smaller percentage of false lexical atoms when the threshold is being relaxed. Such difference may have an impact on the determination of the threshold. Specifically, it may be easier to decide a threshold in the case of PWC measure, since a tighter threshold generally gives a higher precision.

The difference between the WA measure and MI measure seems to be insignificant. The shape of their curves also look very similar. Although the WA and MI measures generally result in lower precision than the PWC measure, the difference is not so significant as it seems to be, because the size of the sample of lexical atoms used for the evaluation is very small. For example, at the cutoff of 30 pairs, the PWC measure only returns 2 more true lexical atoms than the WA or the MI measure. However, there is a great difference between the CS measure and the other three measures. The precision of the CS measure is significantly lower than those of the other three. We have tried different window sizes from 20 to 80 to see if the results from the CS measure might be sensitive to the window size, but there is no significant difference in the results.

It is impressive that the results of the MI measure are comparable with the PWC or WA measure, since this may indicate that other measures for finding collocations may also capture the compositionality (at least to certain extent). Therefore, it makes sense to make a further comparison with other collocation measures such as [Dunning 93], but this is out of the scope of the paper.

Although our experiments have not shown a great difference between the PWC and MI measures, these two measures may capture the compositionality in different ways, since we found that the two measures have ranked different lexical atoms on the top. Each measure found 41 lexical atoms among the top 50 candidates, but the two sets of 41 lexical atoms have only 21 in common. Thus, it is not a surprise that when the results from the PWA and MI measures are merged, the precision is slightly better than either one alone. The merged results, which are shown in Table 4, are obtained by fixing on the top the common candidates (23 altogether) of the top 50 candidates from each measure and then alternatively picking a new pair from each result set.

|              | PWC and MI (merged) |
|--------------|---------------------|
| at 10 pairs  | 0.900               |
| at 20 pairs  | 0.900               |
| at 30 pairs  | 0.900               |
| at 40 pairs  | 0.875               |
| at 50 pairs  | 0.82                |
| at 70 pairs  | 0.8                 |
| at 100 pairs | 0.73                |
| at 150 pairs | 0.653               |
| at 200 pairs | 0.565               |
| at 300 pairs | 0.463               |
| at 400 pairs | 0.380               |
| **Average**  | **0.700**           |

**Table 4. Precision of merged results from PWC and MI measures**

One advantage of the PWC and CS measures over the WA and MI measures is both the PWC and the CS measures also provide an individual compositionality value w.r.t. each word, besides the single combined compositionality value. The individual compositionality measure can give a further breakdown of the compositionality of the pair and can suggest an explanation why the pair is a lexical atom. For example, the PWC values for "heart attack" w.r.t. "heart" and "attack" are 1.016 and 0.056 respectively, which suggests that "attack" does not carry its regular meaning in "heart attack", but "heart" does. However, the two PWC values for "red cross" are 0.023 and 0.015 respectively, which means neither "red" nor "cross" carries its regular meaning in "red cross". In terms of computational complexity, the PWC, MI, and WA measures are significantly better than the CS measure.

An implicit assumption in all these measures is that we assume the corpus is homogeneous so that the use of a candidate phrase is relatively unambiguous and thus the context can reflect the meaning of the phrase well. This assumption is held in tasks such as discovering lexical atoms from text in a particular domain. However, with this assumption, it is impossible to distinguish whether an ambiguous phrase (e.g., white house) is, or is not, a lexical atom in a specific context, since all the measures rely on the global information to certain extent. In this case, the CS has its advantage, since , it might still be possible to measure the "closeness" of the specific context of the phrase and the general context of the constituent words, even though the "specific context" may have only very few information.

## 5. Conclusions

The major contribution of this paper is the introduction of a novel way of exploiting linguistic context for the purpose of identifying lexical atom. We gave a context-dependent definition of lexical atoms in terms of phrase compositionality. We proposed and implemented three different statistical measures to quantitatively measure the uncertainty of phrase compositionality. The experiment results based on such measures have shown that they all can capture phrase compositionality to some extent.

We also found that the mutual information measure is almost as effective as the PWC and WA measures we proposed in terms of capturing non-compositionality. But, the PWC measure has the advantage of providing a breakdown of the compositionality value. With some threshold, which has to be decided empirically, such measures can be practically used to detect lexical atoms from unrestricted text.

Our notion of linguistic context is a very simplified one. The simplification includes: 1). The boundary of the context is limited to certain number of words (e.g., 80 words in the experiments). 2). The content of the context is simply assumed to be a bag of words. Therefore, a lot of useful information such as the syntactic/semantic relations has been ignored. Even the information about the relative position of the words within a context has not considered. 3). The relation between a word and its context is assumed to be statistically modeled. The experiments, however, have shown that even such a simplified notion of context can be very useful.

## Acknowledgements

The author is grateful to Dr. David A. Evans for first pointing out the problem of lexical atoms. Special thanks are due to the three anonymous reviewers, whose comments have improved the paper significantly. The author also wishes to thank Xiang Tong and Yibing Geng for their comments on this work.